\documentclass[letterpaper, 10 pt, conference]{ieeeconf}  

\IEEEoverridecommandlockouts                              

\usepackage{graphics} 
\usepackage{epsfig} 
\usepackage{times} 
\usepackage{amsmath} 
\usepackage{amssymb}  
\usepackage{amsfonts,amssymb}
\usepackage{color}
\usepackage[noadjust]{cite}
\usepackage{caption}
\usepackage{stfloats}
\usepackage{multirow}
\usepackage{subfigure}
\usepackage[hang,flushmargin, symbol]{footmisc}

\usepackage[colorlinks,linkcolor=blue]{hyperref}

\title{\LARGE \bf
\textit{SyreaNet}: A Physically Guided Underwater Image Enhancement Framework Integrating Synthetic and Real Images
}
\author{Junjie Wen$^{1,2}$, Jinqiang Cui$^{2\dag}$, Zhenjun Zhao$^{1,2}$, Ruixin Yan$^{1,2}$, Zhi Gao$^{3}$, Lihua Dou$^{4}$, Ben M. Chen$^{1}$
\thanks{\noindent \footnotesize$^{1}$Department of Mechanical and Automation Engineering, the Chinese University of Hong Kong, Shatin, N.T., Hong Kong.}%
\thanks{\footnotesize$^{2}$Department of Mathematics and Theories, Peng Cheng Laboratory, Shenzhen, China.}
\thanks{\footnotesize$^{3}$School of Remote Sensing and Information Engineering, Wuhan University, Wuhan, China.}
\thanks{\footnotesize$^{4}$School of Automation, Beijing Institute of Technology, Beijing, China.}
\thanks{\footnotesize$^{\dag}$Corresponding author.}
}

\begin{document}

\maketitle
\thispagestyle{empty}
\pagestyle{empty}

\begin{abstract}
Underwater image enhancement (UIE) is vital for high-level vision-related underwater tasks. Although learning-based UIE methods have made remarkable achievements in recent years, it's still challenging for them to consistently deal with various underwater conditions, which could be caused by: 1) the use of the simplified atmospheric image formation model in UIE may result in severe errors; 2) the network trained solely with synthetic images might have difficulty in generalizing well to real underwater images. In this work, we, for the first time, propose a framework \textit{SyreaNet} for UIE that integrates both synthetic and real data under the guidance of the revised underwater image formation model and novel domain adaptation (DA) strategies. First, an underwater image synthesis module based on the revised model is proposed. Then, a physically guided disentangled network is designed to predict the clear images by combining both synthetic and real underwater images. The intra- and inter-domain gaps are abridged by fully exchanging the domain knowledge. Extensive experiments demonstrate the superiority of our framework over other state-of-the-art (SOTA) learning-based UIE methods qualitatively and quantitatively. The code and dataset are publicly available at \href{https://github.com/RockWenJJ/SyreaNet.git}{https://github.com/RockWenJJ/SyreaNet.git}.
\end{abstract}

\section{INTRODUCTION}

The acquisition of clear underwater images plays a critical role in fields where autonomous underwater vehicles (AUV) are required to interact with underwater environments \cite{johnson2017high}\cite{lin2020roimix}\cite{jesus2022underwater}. However, the complex underwater conditions usually lead to serious quality degradation of underwater images. Hence, UIE is vital for vision-related underwater tasks.

UIE has been studied intensively in recent years \cite{drews2013transmission}\cite{li2021underwater}\cite{fu2020underwater}. Most of them are guided by the simplified atmospheric image formation model:
\begin{equation}
    I_c(x) = J_c(x)\cdot e^{-\beta_c\cdot z} + B_c^\infty \cdot (1-e^{-\beta_c\cdot z}), \label{eq1}
\end{equation}
where $c\in\{R, G, B\}$ is the color channel, $I_c(x)$ is the captured underwater image, $J_c(x)$ is the scene radiance at $x$, $e^{-\beta_c\cdot z}$ is the medium transmission, $B_c^\infty$ is the homogeneous background light, $z$ is the range between the camera and the objects. The goal of the UIE is to recover $J_c(x)$ from $I_c(x)$. However, the overwhelming usage of the simplified model in UIE may introduce significant errors \cite{akkaynak2017space}\cite{akkaynak2018revised}. 

\begin{figure}[ht]
    \setlength{\belowcaptionskip}{-0.5cm}
    \centering
    \includegraphics[width=0.48\textwidth]{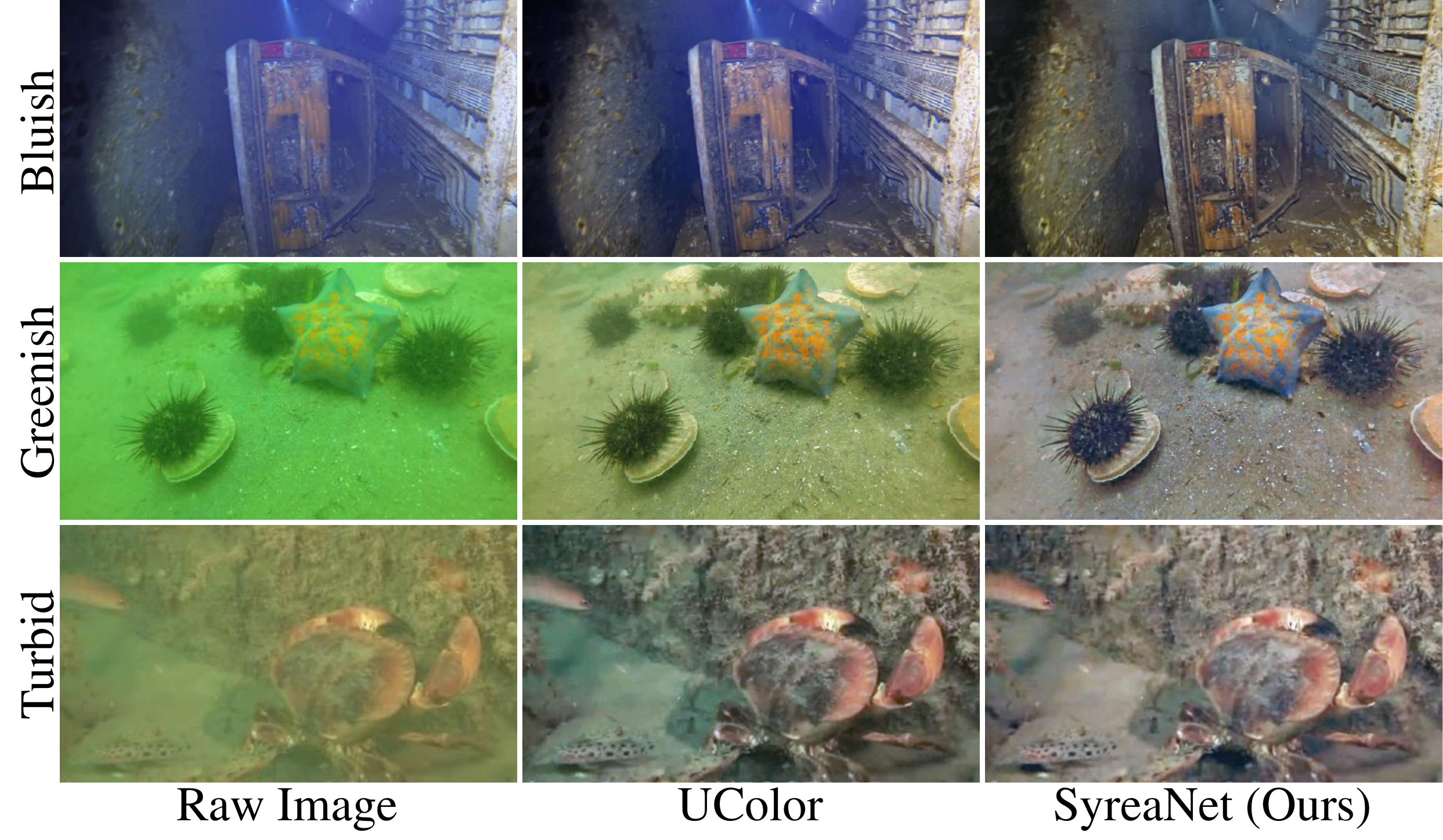}
    \captionsetup{font={small}}
    \caption{Enhancement examples under various underwater conditions. The second and third columns are the results of UColor \cite{li2021underwater} and our proposed \textit{\textbf{SyreaNet}}, respectively\protect\footnotemark .}
    \label{fig1}
\end{figure}

Learning-based UIE methods \cite{li2021underwater}\cite{fu2020underwater} have achieved significant results in recent years. They are usually trained with synthesized images due to the lack of reference real underwater images. However, the inter-domain gap between the synthetic and real underwater images could make them hard to generalize well on real underwater images. To narrow the inter-domain gap, some researchers have attempted to generate more realistic underwater images \cite{hou2020benchmarking}\cite{ye2022underwater}. Other researchers adopted the idea in transfer learning to alleviate the inter-domain difference \cite{liu2021synthetic}\cite{jiang2022two}. Nevertheless, transferring the synthetic domain knowledge with an improved synthetic dataset has not been studied in the UIE field.

\footnotetext[1]{Video can be found at \href{https://youtu.be/DyOktx7_9JQ}{https://youtu.be/DyOktx7\_9JQ}.}
In addition to the inter-domain gap issue, complex underwater environments often lead to intra-domain gap problems. As shown in Fig. \ref{fig1}, the wavelength-dependent light absorption could lead to a dominant bluish/greenish color cast, while the light backscattering and suspended particles could cause a turbid/hazy effect. Although some methods \cite{uplavikar2019all}\cite{berman2017diving} deal with the problem by classifying synthesized underwater images with Jerlov water types \cite{jerlov1977classification}, the intra-domain adaptation using both synthetic and real underwater images has not been explored.



In this work, we propose \textit{\textbf{SyreaNet}}, a UIE framework combining both synthetic and real underwater images guided by the revised underwater image formation model \cite{akkaynak2018revised} and novel DA strategies. Our main contributions are as follows:
\begin{itemize}
    \item Guided by the revised model, a synthesis module to generate a more realistic underwater image dataset from both in-air and real underwater images is proposed. 
    \item A physically disentangled network is well designed to predict both clear underwater images and the critical components in the revised model.
    \item Novel DA strategies are proposed to align various domain gaps by exchanging knowledge across domains.
    \item The superiority of our proposed framework over other SOTA UIE methods is well demonstrated by extensive experiments both qualitatively and quantitatively.
\end{itemize}

The rest of this article is organized as follows. Related work is introduced in Sec. \ref{sec:related_work}. The framework details are presented in Sec. \ref{sec:methodology}. The experiments are shown in Sec. \ref{sec:experiments}. Finally, concluding remarks are drawn in Sec. \ref{sec:conclusions}.


\section{RELATED WORK} \label{sec:related_work}

\subsection{Underwater Image Formation Model}

Unlike the simplified atmospheric image formation model, a more precise revised model \cite{akkaynak2017space}\cite{akkaynak2018revised} is:
\begin{equation}
    I_c(x) = J_c(x)\cdot e^{-\beta_c^D\cdot z} + B_c^\infty \cdot (1-e^{-\beta_c^B\cdot z}), \label{eq2}
\end{equation}
where $\beta_c^D$ is the wideband attenuation coefficient, and $\beta_c^B$ is the backscattering coefficient. Compared to Eq. \ref{eq1}, it is observed that $\beta_c^D\neq\beta_c^B$. Interesting readers could refer to  Akkaynak \textit{et al.} \cite{akkaynak2018revised} for the expressions of $\beta_c^D$ and $\beta_c^B$.


Theoretically, the scene radiance $J_c(x, d)$ of $x$ at underwater depth $d$ can be restored with \cite{akkaynak2019sea}:
\begin{equation}
    J_c(x, d) = (I_c(x) - \hat{B}_c(x))\cdot e^{\hat\beta_c^D(z)z},  \label{eq3}
\end{equation}
where $\hat{B}_c\approx B_c^\infty \cdot (1-e^{-\beta_c^B\cdot z})$ and $\hat{\beta}_c^D \approx \beta_c^D$. However, $J_c(x, d)$ in Eq. \ref{eq4} is not the full restoration of underwater image, since the scene radiance $J_c(x, d)$ at depth $d$ is attenuated by the diffuse downwelling attenuation coefficient $K_d(\lambda)$ \cite{zaneveld1995light} when the light transmits vertically from the sea surface. Assume $J_s$ is the light spectrum above the sea surface, then $J_c$ under the sea surface of depth $d$ is:
\begin{equation}
    J_c = J_s\cdot e^{-K_d(\lambda)d}, \label{eq4}
\end{equation}
where $e^{-K_d(\lambda)d}$ can be estimated by the white point $\hat{W}_c$ of the ambient light at depth $d$ \cite{akkaynak2019sea}. Let $\hat{T}_c = e^{-\hat\beta_c^D(z)z}$, the complete restoration for an underwater image is:
\begin{equation}
    J_s(x) = (I_c(x) - \hat{B}_c(x))/(\hat{T}_c \cdot \hat{W}_c). \label{eq5}
\end{equation}

In this work, the revised underwater image formation model has been fully utilized in both synthesizing underwater images and enhancing degraded underwater images.

\subsection{Underwater Image Enhancement}
UIE methods could be roughly divided into traditional and learning-based methods. Most traditional methods \cite{drews2013transmission}\cite{galdran2015automatic} acquire clear images with prior assumptions based on the simplified model. However, these methods might be invalid in cases where the prior assumptions are not applicable.

Learning-based methods are usually trained with synthetic underwater images, as acquiring clear underwater images is difficult and costly. Although some learning-based UIE methods developed with Generative Adversarial Networks (GANs) \cite{liu2021ipmgan}\cite{du2021unpaired} don't require synthetic images, they only aim to improve the perceptual quality and in some cases may lead to incorrect color restoration. In contrast, synthetic training data provide both degraded and clear reference images, making the trained models more capable of recovering the original color \cite{li2021underwater}\cite{fu2020underwater}. However, they only utilize synthesized datasets and could have difficulty transferring well in real-world underwater images. In this work, we develop an effective learning-based UIE approach using both synthetic and real underwater images.

\begin{figure*}[ht]
    \setlength{\belowcaptionskip}{-0.5cm}
    \centering
    \includegraphics[width=0.92\textwidth]{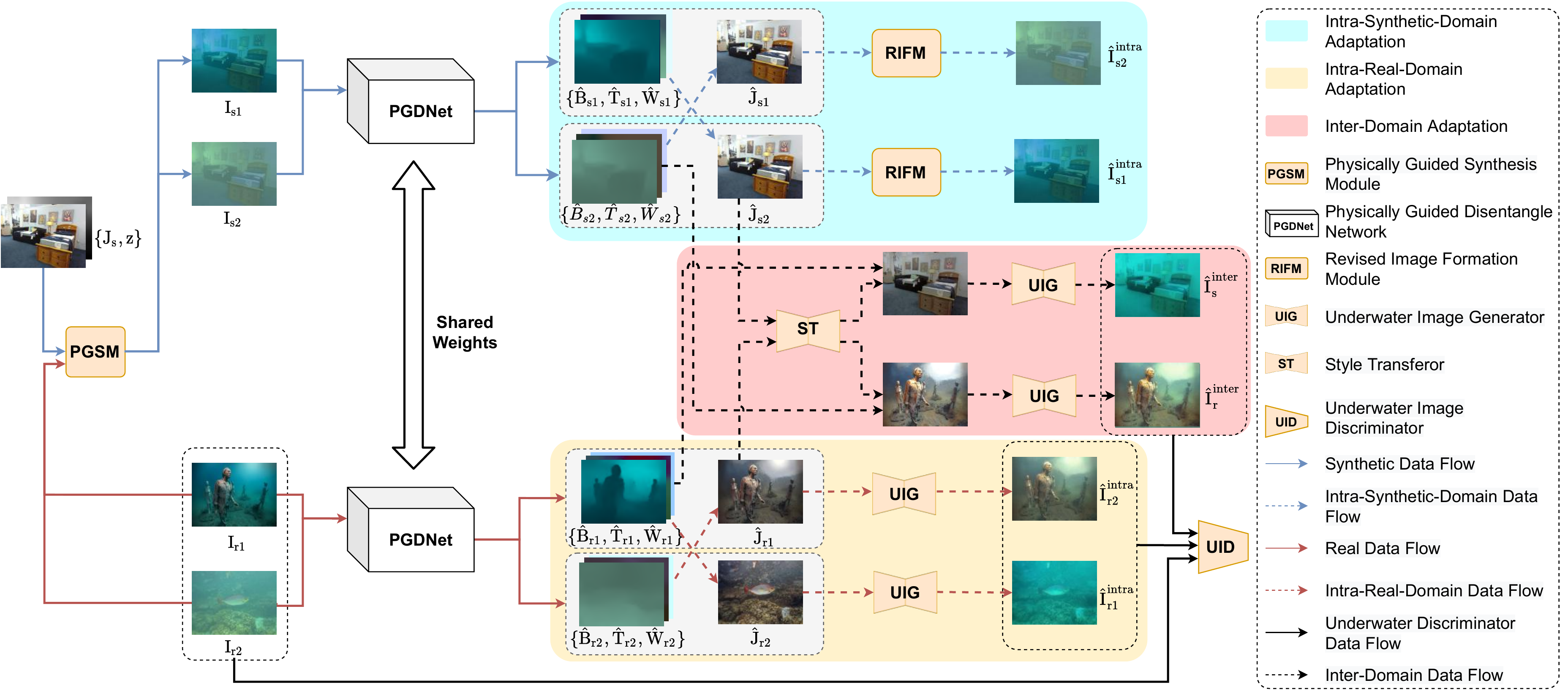}
    \captionsetup{font={small}}
    \caption{Architecture of our proposed UIE framework \textit{\textbf{SyreaNet}}. }
    \label{fig2}
\end{figure*}

\subsection{Underwater Image Synthesis}
Synthesizing underwater images is important for training UIE networks due to the difficulty of obtaining clear real underwater images. Traditional synthesizing methods \cite{hou2020benchmarking}\cite{li2020underwater}\cite{desai2021ruig} usually involve the underwater image formation model. However, they can hardly model complex underwater environments. Moreover, the simplified model they adopted may result in undesired errors.

Other researchers explored learning-based methods to generate realistic underwater images. GANs \cite{li2017watergan}\cite{wang2019uwgan} have been widely adopted for synthesizing images. However, as GANs usually suffer from mode collapse problem \cite{li2021tackling}, the diversity of their synthesized images might be limited. In contrast, Ye \textit{et al.} \cite{ye2022underwater} proposed a neural rendering method with estimated light field maps from real underwater images. But it merely produces synthetic images that look like real underwater images while not conforming to the physical model. In this work, we present a synthesis module to synthesize more realistic underwater images under the guidance of the revised underwater image formation model.

\subsection{Underwater Domain Adaptation}
DA aims to reduce the distribution shift between two different domains. It has been explored in a wide range of tasks \cite{liu2021synthetic}\cite{zheng2018t2net}.

In the UIE field, some studies have already been done to solve the DA problem. Jiang \textit{et al.}\cite{jiang2022two} presented a two-step DA UIE method with CycleGAN \cite{zhu2017unpaired} inspired by transfer learning. Chen \textit{et al.} \cite{chen2022domain} proposed a DA framework via content and style separation in synthesis, real-world and clean domains. Uplavikar \textit{et al.} \cite{uplavikar2019all} proposed a water-type unrelated UIE network with a latent classifier to reduce the domain gaps. However, these methods considered either intra-DA or inter-DA, while the intra-\&inter- DAs have not been well explored. Although Wang \textit{et al.} \cite{wang2021domain} considered both, the intra-DA is only done in the real-world domain, and the dataset is simply classified as hard/easy samples. In this work, we propose a novel intra-\&inter-DA strategy by fully exchanging the domain knowledge.

\section{METHODOLOGY} \label{sec:methodology}

The overall framework of our proposed \textbf{\textit{SyreaNet}} is shown in Fig. \ref{fig2}. Specifically, synthetic underwater images are first generated by our proposed physically guided synthesis module (PGSM) (Sec. \ref{sec:meth_pgs}). Then, various synthetic and real underwater images are fed into the physically guided disentangled network (PGDNet), predicting the clear image $\hat{J}$, backscattering $\hat{B}$, transmission $\hat{T}$ and white point $\hat{W}$ (Sec. \ref{sec:meth_pgdnet}). The intra- and inter-DAs are done by exchanging the knowledge across attribute domains (Sec. \ref{sec:meth_adap}) and training with our well-designed loss functions (Sec. \ref{sec:meth_loss}).

\begin{figure}[hb]
    \setlength{\belowcaptionskip}{-0.5cm}
    \centering
    \includegraphics[width=0.40\textwidth]{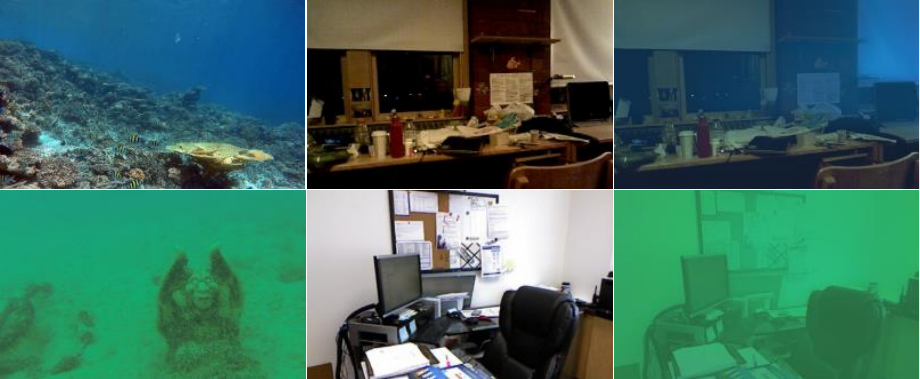}
    \captionsetup{font={small}}
    \caption{Examples of synthetic underwater images. From left to right, the columns are real-world underwater images, in-air images and synthetic underwater images.}
    \label{fig3}
\end{figure}

\subsection{Physically Guided Synthesis Module}\label{sec:meth_pgs}
From Eq. \ref{eq5}, the expression for synthesizing an underwater image is:
\begin{equation}
    I_c(x) = J_s(x) W_c e^{-\hat\beta_c^D(z)z} + \hat{B}_c(x), \label{eq6}
\end{equation}
where $J_s(x)$ is the in-air image, $I_c(x)$ is the synthesized underwater image, and $z$ is the range map.

The in-air image $J_s$ and its range map $z$ are usually easy to get ($z$ could be estimated with depth estimation algorithms \cite{MegaDepthLi18}\cite{monodepth2} if it is not provided). We follow the procedures proposed by Akkaynak \textit{et al.} \cite{akkaynak2019sea} to estimate $\hat{B}_c$, $\hat\beta_c^D$ and $W_c$ from real underwater images. With Eq. \ref{eq6}, we could simulate more realistic synthetic underwater images based on various underwater conditions (Fig. \ref{fig3}).

As shown in Fig. \ref{fig2}, synthetic underwater images $I_{si}$ with the same clear backgrounds but different $\{\hat{B}, \hat{T}, \hat{W}\}$ could be generated through PGSM using in-air image $J_s$, its corresponding range map $z$, and the estimated coefficients from real underwater images $I_{ri}$.

\subsection{Physically Guided Disentangled Network for Underwater Image Enhancement}\label{sec:meth_pgdnet}
\begin{figure}[!h]
    \centering
    \includegraphics[width=0.45\textwidth]{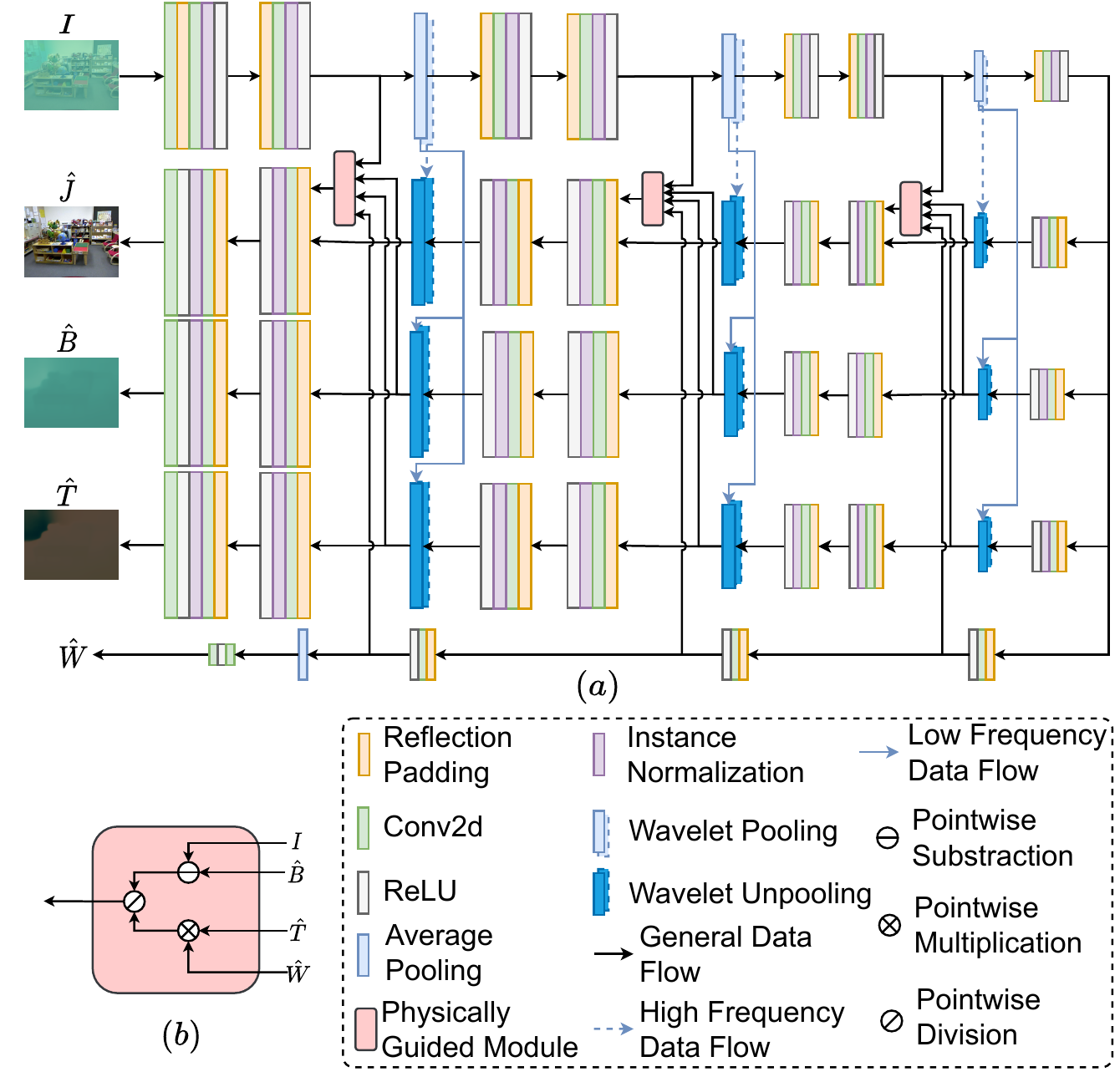}
    \captionsetup{font={small}}
    \caption{The illustration of our proposed PGDNet. (a) The detailed network architecture. (b) The illustration of Physically Guided Module.} 
    \label{fig4}
\end{figure}

The illustrated architecture of our proposed PGDNet is shown in Fig. \ref{fig4}. Specifically, given an underwater image $I$, feature maps of different spatial resolutions are first extracted by an encoder. Then, the clear image $\hat{J}$, the backscattering $\hat{B}$ and the transmission map $\hat{T}$ are generated with three branches of decoders, and the white point $\hat{W}$ is regressed with a branch of fully convolutional network.

For underwater images, it is observed that the color cast and luminance are mainly in low frequency while the details like texture and edge appear at high frequency \cite{huo2021efficient}. We then utilize the wavelet transform (WT) \cite{yoo2019photorealistic} to decompose low- and high-frequency signals. The wavelet pooling has four kernels and therefore outputs four channels \{LL, LH, HL, HH\}, in which LL relates to low-frequency signals like color and luminance while other channels capture high-frequency edge and texture. 

As shown in Fig. \ref{fig4}, to better remove the color cast and improve the details of underwater images, only the high-frequency parts after the wavelet pooling are skip-connected to the wavelet unpooling in the clear image $\hat{J}$ decoder branch. In contrast, the $\hat{B}$ and $\hat{T}$ decoders are only connected with LL as they are dominated by low-frequency signals. 

Moreover, a Physically Guided Module (PGM) based on the revised model (Eq. \ref{eq5}) is introduced on different scales to guide the prediction of the clear image $\hat{J}$.



\subsection{Intra- and Inter-Domain Adaptation}\label{sec:meth_adap}
The core idea of our proposed DA method lies in fully transferring knowledge across domains after getting clear images and their disentangled components.

As shown in Fig. \ref{fig2}, the intra-DA includes intra-synthetic- and intra-real-DAs. In both domains, it is observed that the clear backgrounds $\hat{J}$ conserve similar distributions while the disentangled components \{$\hat{B}, \hat{T}, \hat{W}$\} have various distributions. Therefore, the clear backgrounds are kept invariant while the disentangled components' knowledge is exchanged. The revised image formation module (RIFM) based on Eq. \ref{eq6} is introduced for intra-synthetic-DA to formulate synthetic underwater images directly, as the clear backgrounds and transferred components could be matched. In contrast, for intra-real-DA, the clear underwater backgrounds and exchanged components are recombined with an underwater image generator (UIG) to formulate $\hat{I}^{\text{intra}}_{ri}$. 

For inter-DA, it is noted that both the clear backgrounds and the disentangled components lie in distinct distributions. Therefore, the clear background knowledge between synthetic and real domains is first exchanged with a style transferor (ST). Then the exchanged clear backgrounds and the disentangled components from the other domain are fed into UIG to generate synthetic underwater images.


\subsection{Loss Design}\label{sec:meth_loss}
The total loss of our proposed \textit{\textbf{SyreaNet}} during training is combined with three types of losses and their corresponding balanced weights, \textit{i.e.}, the reconstruction loss $\mathcal{L}_{rec}$, the consistency loss $\mathcal{L}_{con}$ and the adversarial loss $\mathcal{L}_{adv}$:
\begin{equation}
    \mathcal{L}=\lambda_{rec}\mathcal{L}_{rec} + \lambda_{con}\mathcal{L}_{con} + \lambda_{adv}\mathcal{L}_{adv}. \label{eq7}
\end{equation}

\textit{Reconstruction Loss}: The reconstruction loss is designed to guide the predictions of clear image $\hat{J}_{si}$ and the disentangled components $\{\hat{B}_{si}, \hat{T}_{si}, \hat{W}_{si}\}$ of synthetic images in a supervised manner. Thus it is defined as:
\begin{equation}
    \begin{aligned}
    \mathcal{L}_{rec}= & \sum_{i}(||J_{si} - \hat{J}_{si}||_{2} + ||B_{si} - \hat{B}_{si}||_{2} + ||T_{si} - \hat{T}_{si}||_{2}\\
                       &  + ||W_{si} - \hat{W}_{si}||_{2} + \mathcal{L}_{ssim}(J_{si}, \hat{J}_{si})), \label{eq8}
    \end{aligned}
\end{equation}
where $\mathcal{L}_{ssim}(J_{si}, \hat{J}_{si})$ denotes the structural similarity loss \cite{wang2004image} between $J_{si}$ and $\hat{J}_{si}$.

\textit{Consistency Loss}: Consistency loss is introduced to coordinate the intra-synthetic-domain data distributions. First, it is noted that the predicted clear in-air images $\hat{J}_{si}, i\in \{ 1,2\}$ with the same background should be consistent with each other. Then, the regenerated synthetic underwater images $\hat{I}^{intra}_{si}$ should be consistent with the original synthetic images $I_{si}$. Therefore, the consistency loss $\mathcal{L}_{con}$ is defined as:
\begin{equation}
\begin{aligned}
\mathcal{L}_{con} = & ||\hat{J}_{s1} - \hat{J}_{s2}||_{2} + \sum_{i}||I_{si}-\hat{I}^{\text{intra}}_{si}||_{2}. \label{eq10}
\end{aligned}
\end{equation}

\textit{Adversarial Loss}: Adversarial loss is designed to align the intra-real-domain and inter-domain knowledge. As shown in Fig. \ref{fig2}, the UIG first tries to fool the UID by generating more realistic underwater images from the exchanged data in intra-real- and inter-domains, while the UID attempts to distinguish the regenerated underwater images from real underwater images. Thus the adversarial loss is defined as:
\begin{equation}
\begin{aligned}
\mathcal{L}_{adv} = & \mathbb{E}_{\hat{I}^{\text{inter}}}[\log(1-D(\hat{I}^{\text{inter}})] + \\ 
                    & \mathbb{E}_{\hat{I}^{\text{intra}}_r}[\log(1-D(\hat{I}^{\text{intra}}_r)] + \mathbb{E}_{I_r}[\log(D(I_r)]. \label{eq11}
\end{aligned}
\end{equation}

\section{EXPERIMENTS} \label{sec:experiments}
In this section, we first introduce the experiment setups (Sec. \ref{sec:exp_implement_details}). Then we compare our proposed PGSM with other underwater synthesis methods (Sec. \ref{sec:exp_syn}). The enhancement results of our framework are validated both qualitatively (Sec. \ref{sec:exp_quali}) and quantitatively (Sec. \ref{sec:exp_quan}). Finally, ablation studies are shown in Sec. \ref{sec:exp_ab}.

\begin{figure*}[t]
    \centering
    \includegraphics[width=0.95\textwidth]{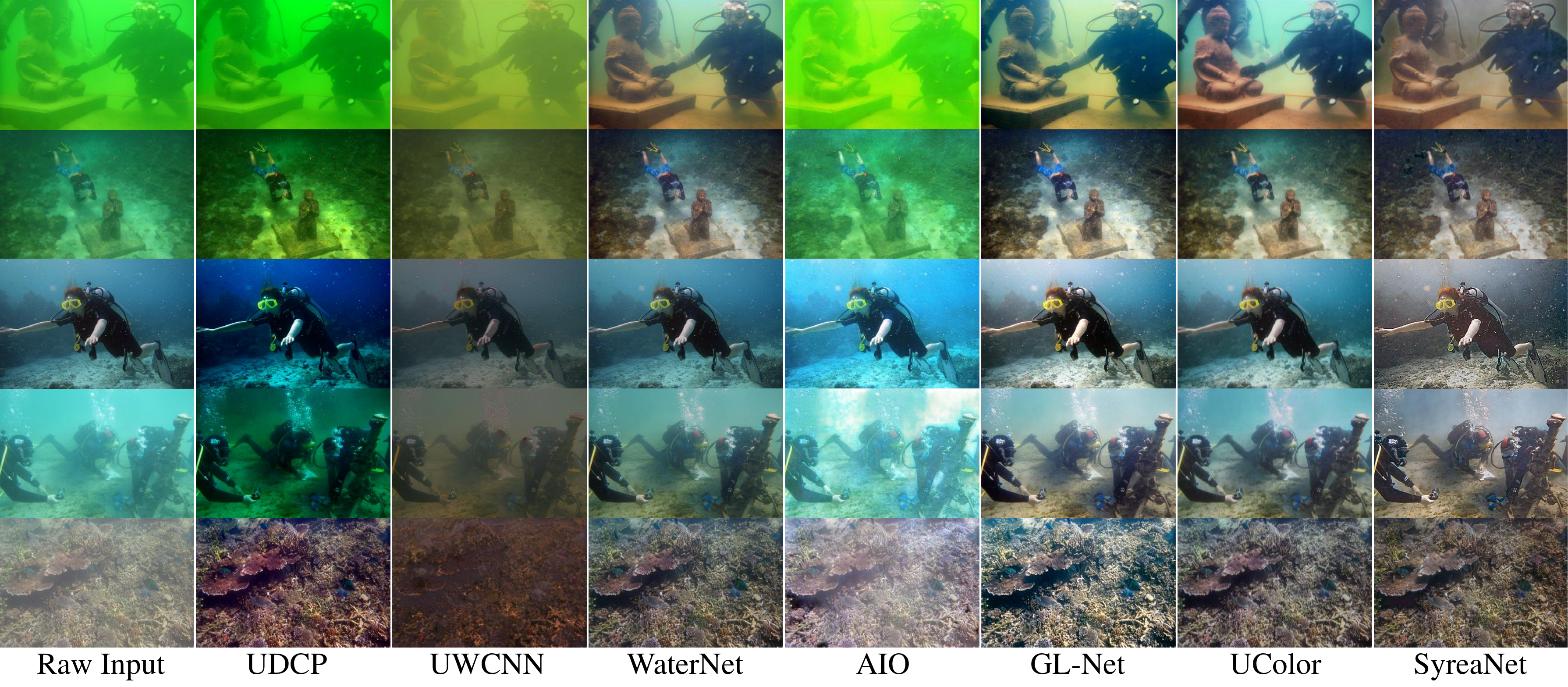}
    \captionsetup{font={small}}
    \caption{Qualitative comparison of the performances of \textit{\textbf{SyreaNet}} and other SOTA UIE methods on real underwater images.}
    
    \label{fig6}
\end{figure*}

\subsection{Implementation Details} \label{sec:exp_implement_details}
Our proposed \textit{\textbf{SyreaNet}} is trained with PyTorch on a single NVIDIA Tesla V100 GPU. The style transferor $\text{WCT}^2$ \cite{yoo2019photorealistic} is utilized for transferring inter-domain background knowledge, and the discriminator of PatchGAN \cite{isola2017image} is employed as the UID. The training images are resized to $256\times 256$ with a batch size of 12. The balanced weights $\lambda_{rec}, \lambda_{con}$ and $\lambda_{adv}$ are set as 10, 1, and 2, respectively. Due to the difficulty of training the whole framework from scratch, the initial weights of PGDNet are pretrained by synthetic images without any DA strategies for 100 epochs with Adam optimizer of learning rate 1e-3. Then the whole framework is finetuned by both synthetic and real underwater images for 100 epochs with a learning rate of 1e-6. The synthesized images are based on 20k randomly selected in-air images from NYUDepth \cite{Silberman:ECCV12}, SUNRGBD \cite{zhou2014learning}, KITTI \cite{geiger2012we} and MegaDepth \cite{MegaDepthLi18}. There are 5k real underwater images randomly selected from UIEB \cite{li2019underwater} and EUVP \cite{islam2020fast} for training.


\subsection{Comparison of Synthetic Underwater Images} \label{sec:exp_syn}
\begin{figure}[hb]
    \setlength{\belowcaptionskip}{-0.5cm}
    \centering
    \includegraphics[width=0.35\textwidth]{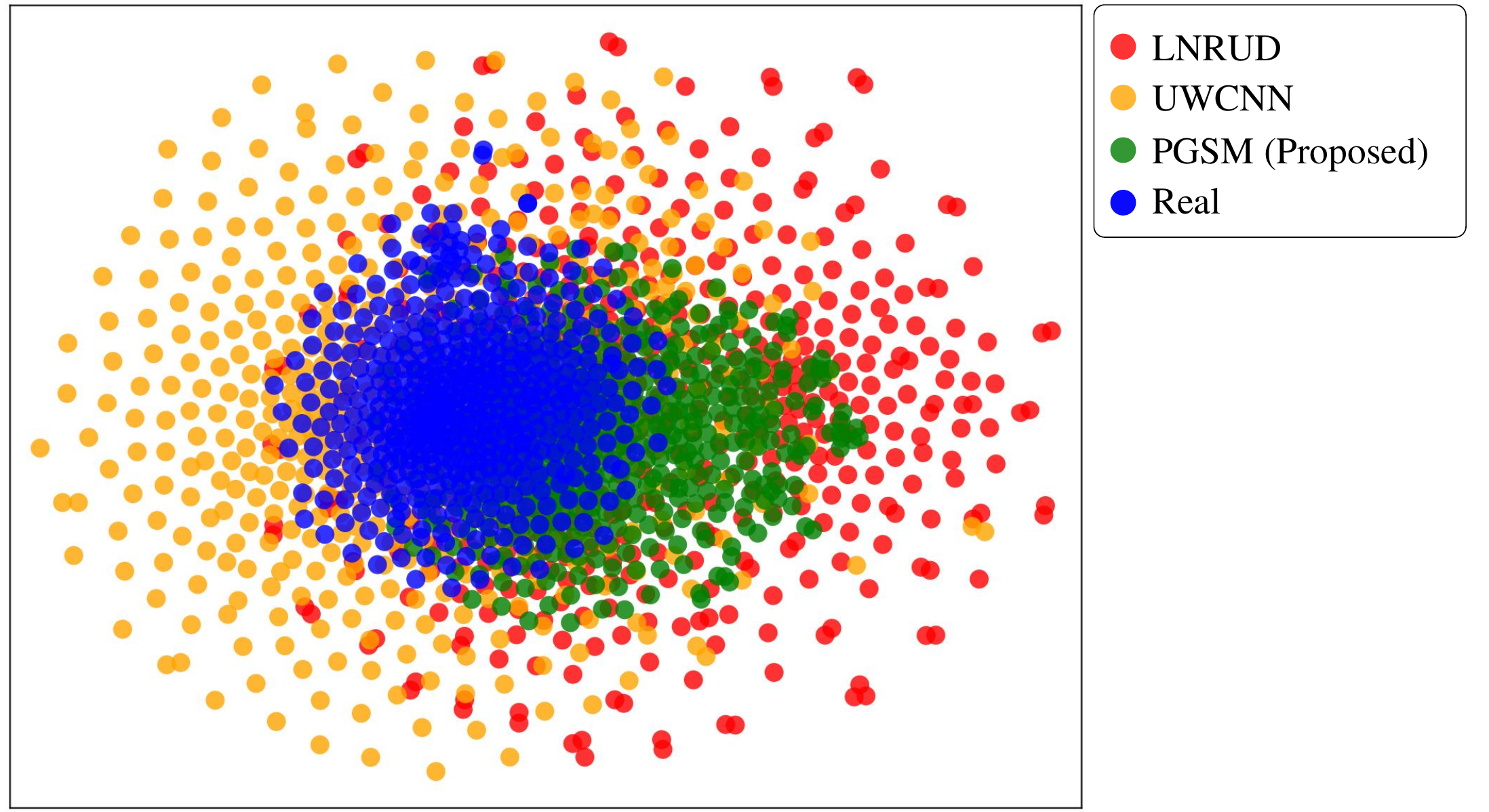}
    \captionsetup{font={small}}
    \caption{Visualization results of t-SNE on different synthesized underwater dataset and real underwater dataset.}
    \label{fig5}
\end{figure}

To validate the superiority of our proposed PGSM, we visualize the t-SNE \cite{van2008visualizing} results of both real-world underwater images and synthesized underwater images from LNRUD \cite{ye2022underwater}, UWCNN \cite{li2020underwater} and PGSM. As shown in Fig. \ref{fig5}, the synthesized images from PGSM have the largest overlapped region with the real dataset. To quantitatively measure the similarity, we further compute the Intersection Ratio (IR) and the Center Distance (CD) of the clustered t-SNE points of the synthesized datasets to the real dataset. As shown in Tab. \ref{tab:dataset}, it demonstrates that the synthesized images from PGSM have the smallest domain gap with real underwater images compared to other synthesized images.

\begin{table}[h]
  \centering
  \captionsetup{font={small}}
  \caption{Intersection Ratio (IR) and Center Distance (CD) to Real Underwater Images.}
  \label{tab:dataset}
  \setlength{\tabcolsep}{1.5mm}{
  \begin{tabular}{c|ccc}
    \hline  
    Method & LNRUD & UWCNN & Proposed \\ 
    \hline  
    IR$\uparrow$ & 35.4\% & 40.9\% & \textbf{61.3\%}  \\ 
    \hline
    CD$\downarrow$ & 133.96 &  96.82 & \textbf{36.14} \\
    \hline
    \end{tabular}
  }
\vspace{-1em}
\end{table}

\subsection{Qualitative Comparison of UIE methods} \label{sec:exp_quali}
We qualitatively compare the performance of our proposed \textit{\textbf{SyreaNet}} with other SOTA methods on real underwater images. The results from other methods are either implemented by the original codes or provided by the authors. Thoese methods include traditional method UDCP \cite{drews2013transmission} and learning-based methods UWCNN \cite{li2020underwater}, WaterNet \cite{li2019underwater}, AIO \cite{uplavikar2019all}, GL-Net \cite{fu2020underwater} and UColor \cite{li2021underwater}.

As shown in Fig. \ref{fig6}, it can be shown that although UDCP \cite{drews2013transmission} is effective in erasing the hazy effect, it fails to deal with bluish/greenish underwater conditions and also results in a dark tone. UWCNN \cite{li2020underwater} fails to enhance underwater images effectively, producing a reddish and dark tone. AIO \cite{uplavikar2019all} brightens the images but introduces obvious artifacts. WaterNet \cite{li2019underwater} is not effective in bluish environments (the 3rd and 4th row). Although GL-Net \cite{fu2020underwater} and UColor \cite{li2021underwater} can enhance underwater images in a variety of underwater conditions, they still fail to effectively remove the bluish/greenish effect for relatively distant objects. In contrast, our proposed \textit{\textbf{SyeraNet}} outperforms other SOTA methods in enhancing the degraded images in various underwater conditions by consistently removing the water effect (bluish/greenish/hazy).

\begin{figure}[htb]
    \setlength{\belowcaptionskip}{-0.5cm}
    \centering
    \includegraphics[width=0.48\textwidth]{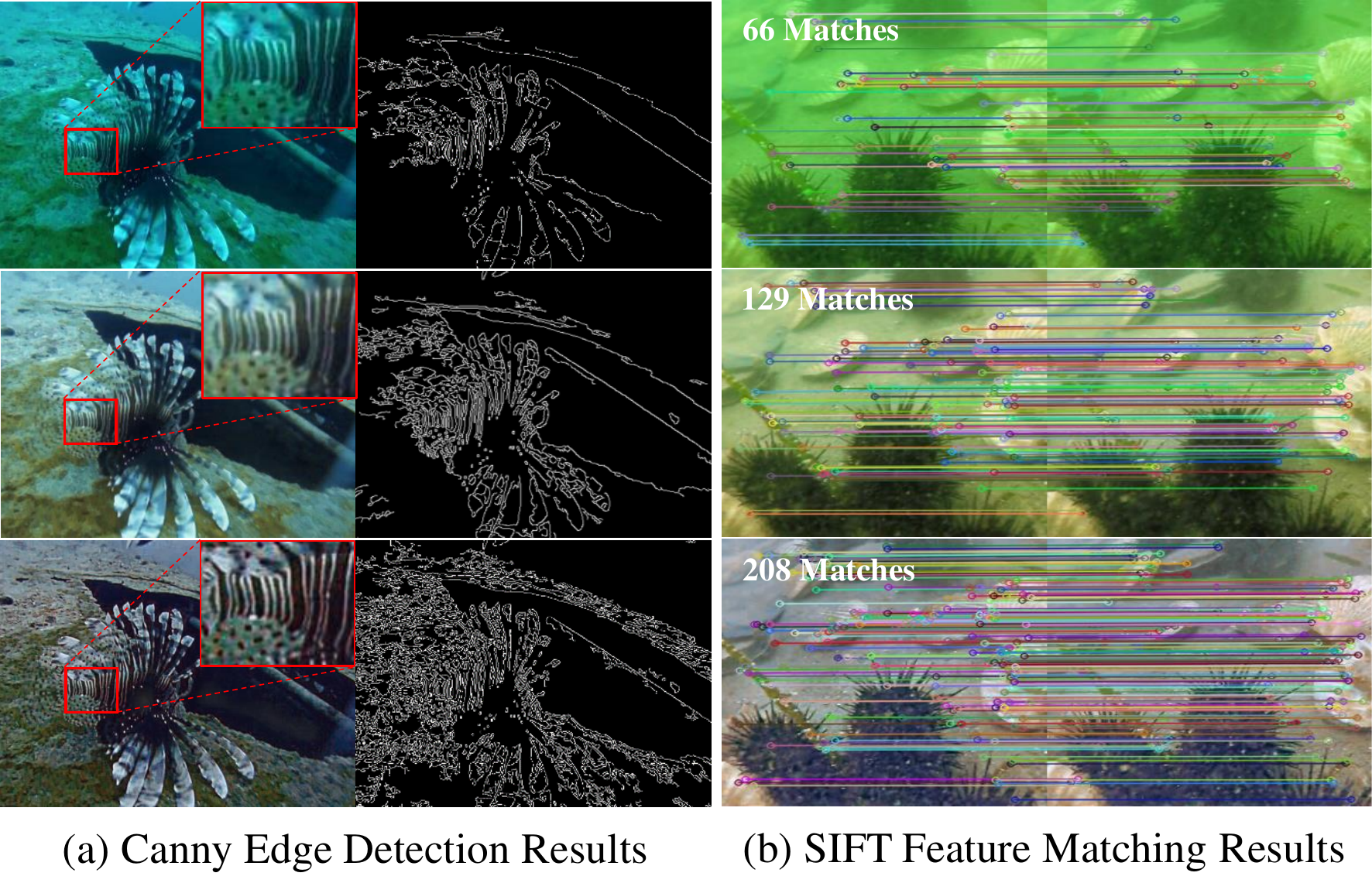}
    \captionsetup{font={small}}
    \caption{Visualization example of canny edge detection and SIFT feature matching results of the raw underwater image (1st row), UColor (2nd row) and \textit{\textbf{SyreaNet}} (3rd row).}
    \label{fig7}
\end{figure}

In addition to the effectiveness in removing the water effect, \textit{\textbf{SyreaNet}} is also superior in restoring the details of degraded underwater images, as shown in Fig. \ref{fig7}. 

\subsection{Quantitative Comparison} \label{sec:exp_quan}
\begin{figure}[htb]
    \setlength{\belowcaptionskip}{-0.5cm}
    \centering
    \includegraphics[width=0.5\textwidth]{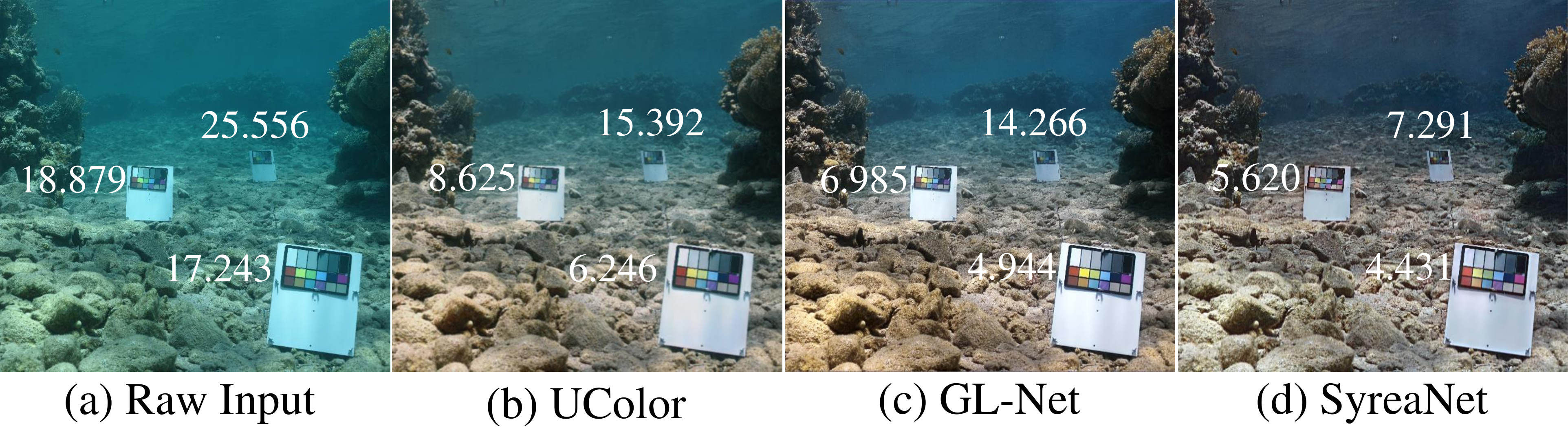}
    \captionsetup{font={small}}
    \caption{Example of the RGB error (the lower the better) of various UIE methods.}
    \label{fig8}
\end{figure}

We further compute the non-reference metrics UIQM \cite{panetta2015human} and UCIQE \cite{yang2015underwater}, which are commonly used for evaluating underwater image qualities, to compare the performance of various UIE methods quantitatively. The evaluated images are selected from UIEB \cite{li2019underwater} and EUVP \cite{islam2020fast} datasets and are not used for training. As shown in Tab. \ref{tab:uiqm_uciqe}, our proposed \textit{\textbf{SyreaNet}} achieved comparable performances with other SOTA UIE methods in both datasets with respect to UIQM/UCIQE. However, as these two metrics are only designed to measure the perceptual quality with heuristic assumptions and have some limitations \cite{li2018emerging}, we compute the average RGB error \cite{akkaynak2019sea}\cite{berman2020underwater} with the representative images in SQUID \cite{berman2020underwater} to evaluate the color restoration capability. As shown in Fig. \ref{fig8}, our \textit{\textbf{SyreaNet}} is superior to other methods in restoring the original color, especially for distant objects. The last row in Tab. \ref{tab:uiqm_uciqe} also demonstrates that our proposed method outperforms other SOTA methods by a large margin.

\begin{table}[h]
  \centering
  \captionsetup{font={small}}
  \caption{Quantitative comparison of our proposed method and other SOTA methods}
  \label{tab:uiqm_uciqe}
  \setlength{\tabcolsep}{1mm}{
  \begin{tabular}{c|c|c|c|c|c|c}
    \hline  
    \multicolumn{2}{c|}{Methods} &AIO & GLNet & WaterNet & UColor & SyreaNet\\ 
    \hline  
    \multirow{2}{*}{UIQM$\uparrow$} & UIEB & 1.069 & \textit{1.598} & 1.315 & 1.372 & \textbf{1.656}  \\ 
    \cline{2-7}
    & EUVP & 1.021 & \textit{1.359} & 1.213 & 1.247 & \textbf{1.363} \\
    \hline
    \multirow{2}{*}{UCIQE$\uparrow$} & UIEB & 0.506 & \textbf{0.619} & 0.543 & 0.558 & \textit{0.582} \\ 
    \cline{2-7}
    & EUVP & 0.497 & \textbf{0.614} & 0.523 & 0.541 & \textit{0.578} \\
    \hline 
    \multicolumn{2}{c|}{RGB Error$\downarrow$} & 26.574 & \textit{10.488} & 12.167 & 11.332 & \textbf{5.158} \\ 
    \hline 
    \end{tabular}
  }
\vspace{-1em}
\end{table}

\subsection{Ablation Study} \label{sec:exp_ab}
In this section, we investigate the effectiveness of each module in our proposed \textit{\textbf{SyreaNet}}. The effectiveness is compared on both synthetic and real underwater datasets. The metrics for synthetic images are PSNR and SSIM, while the metric for real underwater images is the RGB error.

\begin{table}[h]
  \centering
  \captionsetup{font={small}}
  \caption{Ablation studies on synthetic and real data.}
  \label{tab:ablation}
  \setlength{\tabcolsep}{1.5mm}{
  \begin{tabular}{c|cc|c}
    \hline  
    
    \hline
    Ablations & PSNR$\uparrow$ & SSIM$\uparrow$ & RGB Error$\downarrow$ \\
    \hline  
    
    \hline
    w/o PGSM & \textbf{25.49} & \textbf{0.947} & 8.936   \\
    \hline
    w/o WT & 22.78 & 0.933 & 5.819 \\
    w/o PGM & 22.49 & 0.930 & 5.566 \\
    \hline
    w/o Intra-Real-DA & 24.81 & 0.938 & 6.245 \\
    w/o Intra-Synthetic-DA & 22.23 & 0.925 & 6.529 \\
    w/o Inter-DA & 24.51 & 0.932 & 6.428 \\
    \hline
    
    \hline
    Proposed &  24.92 & 0.945 & \textbf{5.158}  \\
    \hline 
    
    \hline
    \end{tabular}
    }
\end{table}

\subsubsection{Effectiveness of Physically Guided Synthesis Module}
We investigate the influence of PGSM by replacing it with the synthesis method \cite{li2020underwater} based on the simplified model. As shown in the first row of Tab. \ref{tab:ablation}, while PSNR and SSIM are marginally better, possibly as a result of synthesized images based on simplified model being easier for the network to train, the color restoration ability has dramatically declined, highlighting the significance of our proposed PGSM in effectively enhancing real-world underwater images. The visualization result in Fig. \ref{fig10} shows that the framework without PGSM could cause inaccurate color restoration (in this example, cause a reddish tone).




\subsubsection{Effectiveness of Physically Guided Disentangled Network}
We also test the effectiveness of WT and PGM in our PGDNet. The performance drop in Tab. \ref{tab:ablation} clearly shows the benefits brought by the two modules. As shown in Fig. \ref{fig10}, the result in the absence of WT shows difficulty in dealing with image details and uniformly removing the water effect.

\subsubsection{Effectiveness of Domain Adaptation Strategies}
\begin{figure}[htb]
\setlength{\abovecaptionskip}{-0.1cm}
\setlength{\belowcaptionskip}{-0.5cm}
\centering
\subfigure[Before Inter-DA]{
\begin{minipage}[t]{0.46\linewidth}
\centering
\includegraphics[width=3.15cm]{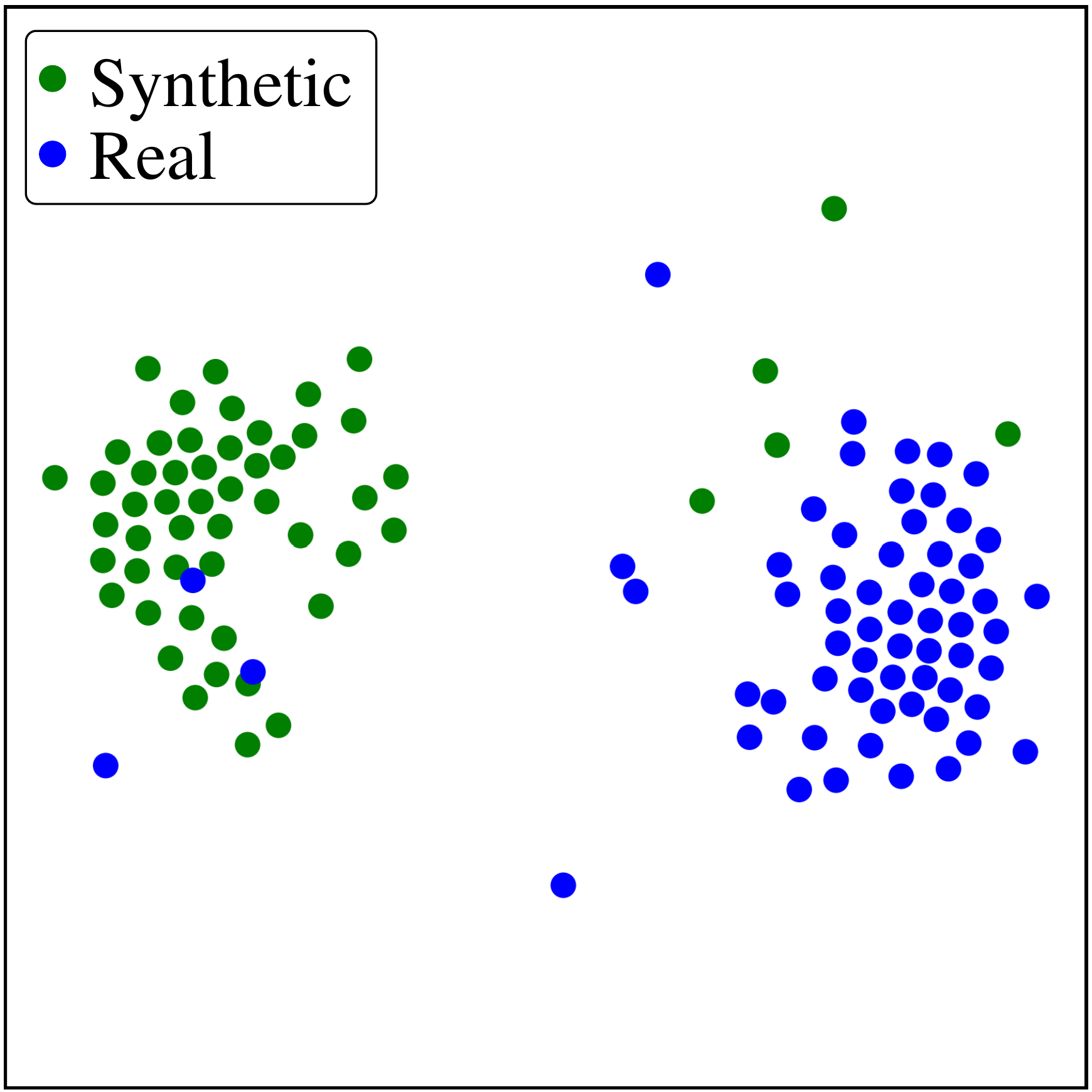}
\end{minipage}
}
\subfigure[After Inter-DA]{
\begin{minipage}[t]{0.46\linewidth}
\centering
\includegraphics[width=3.15cm]{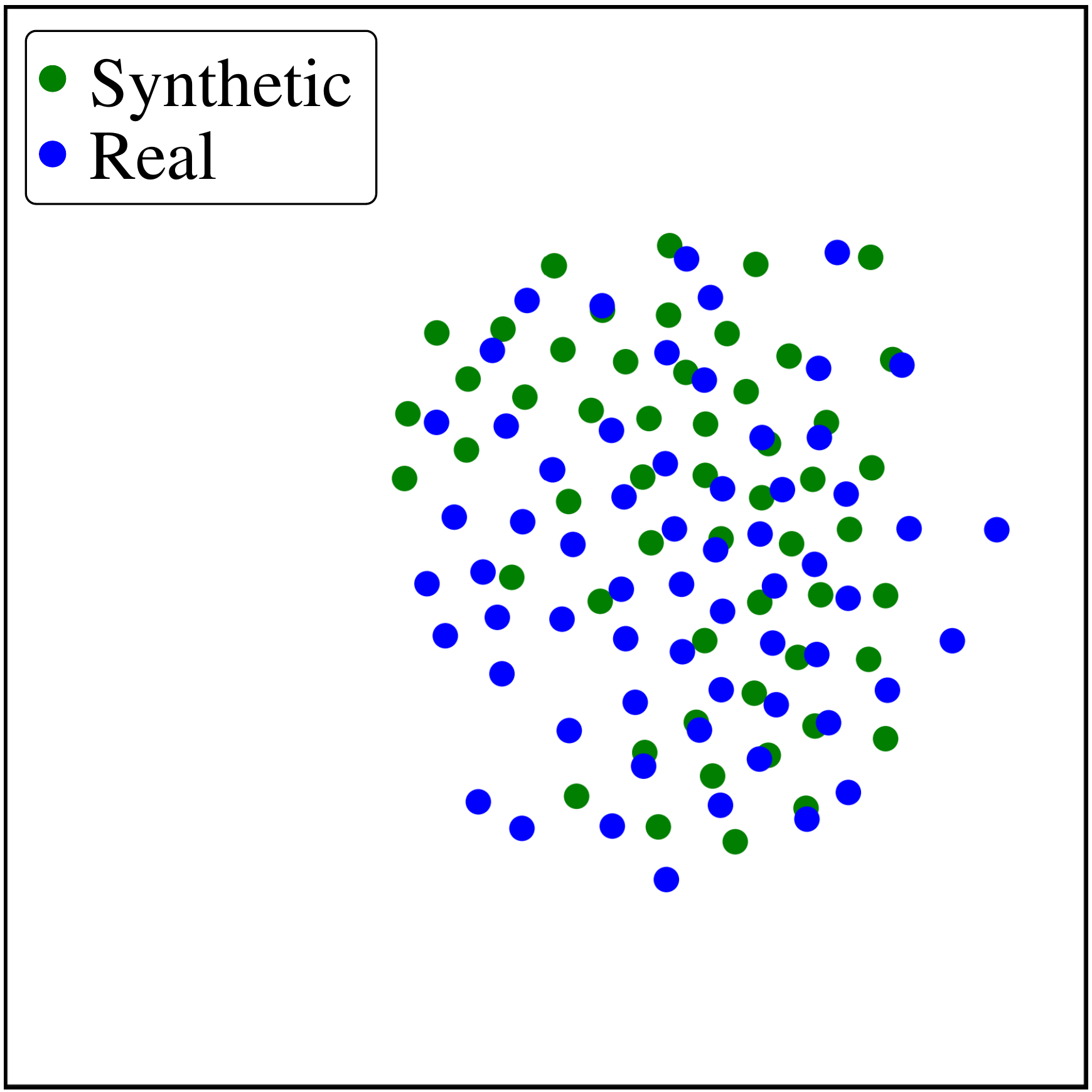}
\end{minipage}
}
\captionsetup{font={small}}
\caption{Visualization of t-SNE results on encoded feature of synthetic and real underwater images before and after inter-DA.}
\label{fig9}
\end{figure}

\begin{figure}[htb]
    \setlength{\belowcaptionskip}{-0.5cm}
    \centering
    \includegraphics[width=0.48\textwidth]{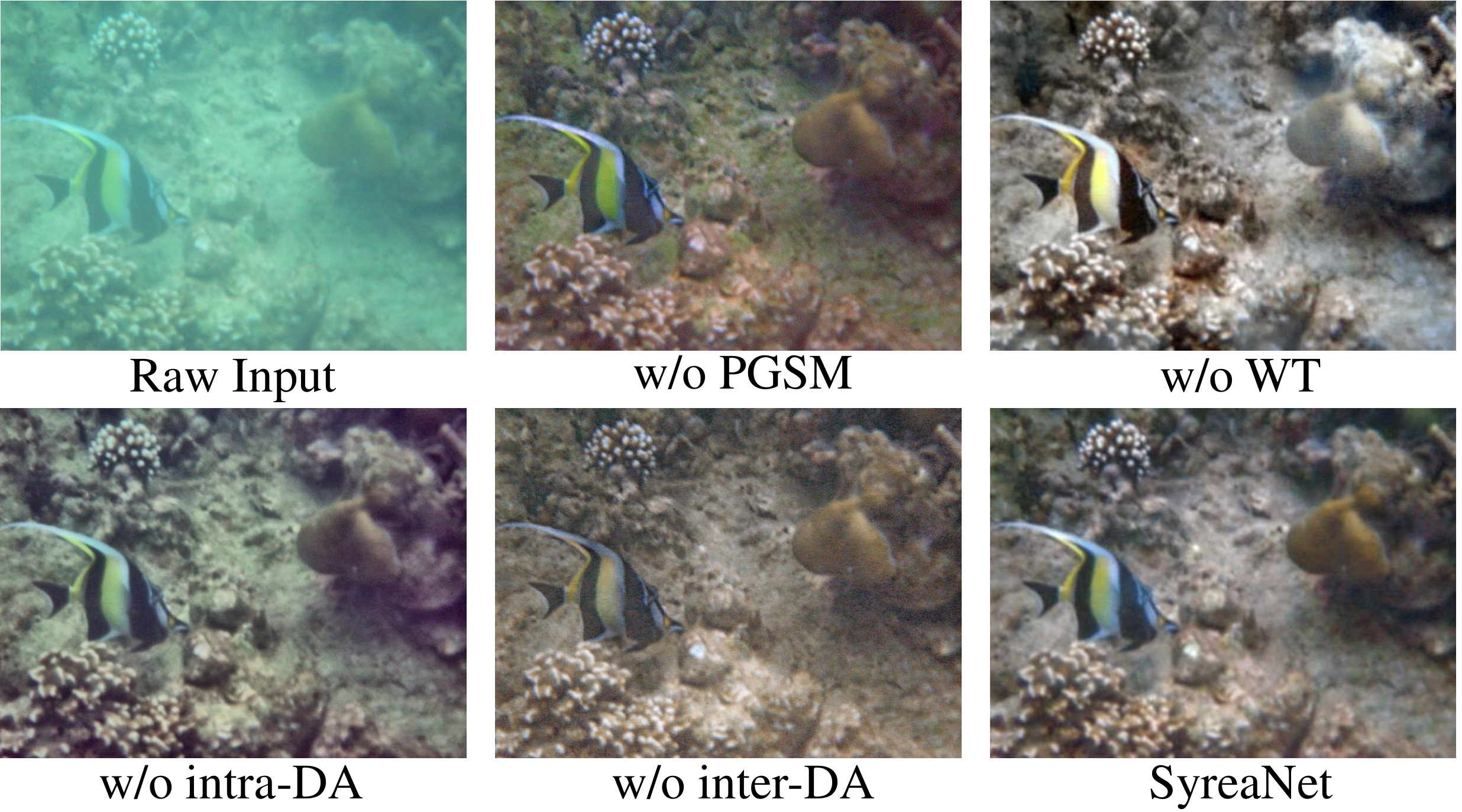}
    \captionsetup{font={small}}
    \caption{Example of UIE results with different ablations.}
    \label{fig10}
\end{figure}

We further investigate the effectiveness of our proposed intra-synthetic-DA, intra-real-DA, and inter-DA. As shown in Tab. \ref{tab:ablation}, it is observed that intra-synthetic-DA has a larger impact than other DA modules since the consistency loss for aligning intra-synthetic-domain data distributions serves as a strong constraint in the training process. Fig. \ref{fig9} also clearly shows the effectiveness of inter-DA in coordinating the inter-domain data distributions. As shown in Fig. \ref{fig10}, the result without intra-DA seems ineffective in dealing with various underwater conditions (still greenish in the UIE result). In contrast, the framework without inter-DA could result in a relatively dark tone.



\section{CONCLUSIONS} \label{sec:conclusions}
In this study, we have proposed a novel UIE framework \textit{\textbf{SyreaNet}} that combines synthetic and real data under the guidance of the revised underwater image formation model and DA strategies. First, we have proposed PGSM to generate synthesized underwater images that have smaller domain gap with real images than other synthesis methods. The well-designed PGDNet has also been presented to predict the disentangled components and clear images with better color and detail restoration capability. Novel intra- and inter-DA strategies that fully exchange domain knowledge have been proposed based on the observation that intra-domain images conserve similar background distributions while the inter-domain images do not. Extensive experiments indicate that our framework, which outperforms previous SOTA learning-based UIE approaches in restoring the original color and details of degraded underwater images, could lay a sound basis for future vision-related AUV tasks.





\bibliographystyle{ieeetr}
\bibliography{IEEEexample}

\end{document}